\documentclass[10pt,twocolumn,letterpaper]{article}

\usepackage{iccv}
\usepackage{times}
\usepackage{epsfig}
\usepackage{graphicx}
\usepackage{amsmath}
\usepackage{amssymb}
\usepackage{subcaption}
\usepackage{multirow}
\usepackage{authblk}

\usepackage[breaklinks=true,bookmarks=false]{hyperref}

\iccvfinalcopy 

\ificcvfinal\fi

\begin{document}

\title{Aria Digital Twin: A New Benchmark Dataset for Egocentric 3D Machine Perception}

\author[1]{Xiaqing Pan}
\author[1]{Nicholas Charron}
\author[1]{Yongqian Yang}
\author[1]{Scott Peters}
\author[1]{Thomas Whelan}
\author[1]{Chen Kong}
\author[1]{Omkar Parkhi}
\author[1]{Richard Newcombe}
\author[1]{Carl Yuheng Ren}
\affil[1]{Meta Reality Labs \authorcr\{\tt xiaqingp, nickcharron, yongqian, scpeters, twhelan, chenk, omkar, newcombe, carlren\}@meta.com}

\maketitle
\ificcvfinal\thispagestyle{empty}\fi
\graphicspath{ {./images/} }

\begin{figure*}
\begin{center}
  \begin{subfigure}[b]{\textwidth}
    \includegraphics[width=\textwidth]{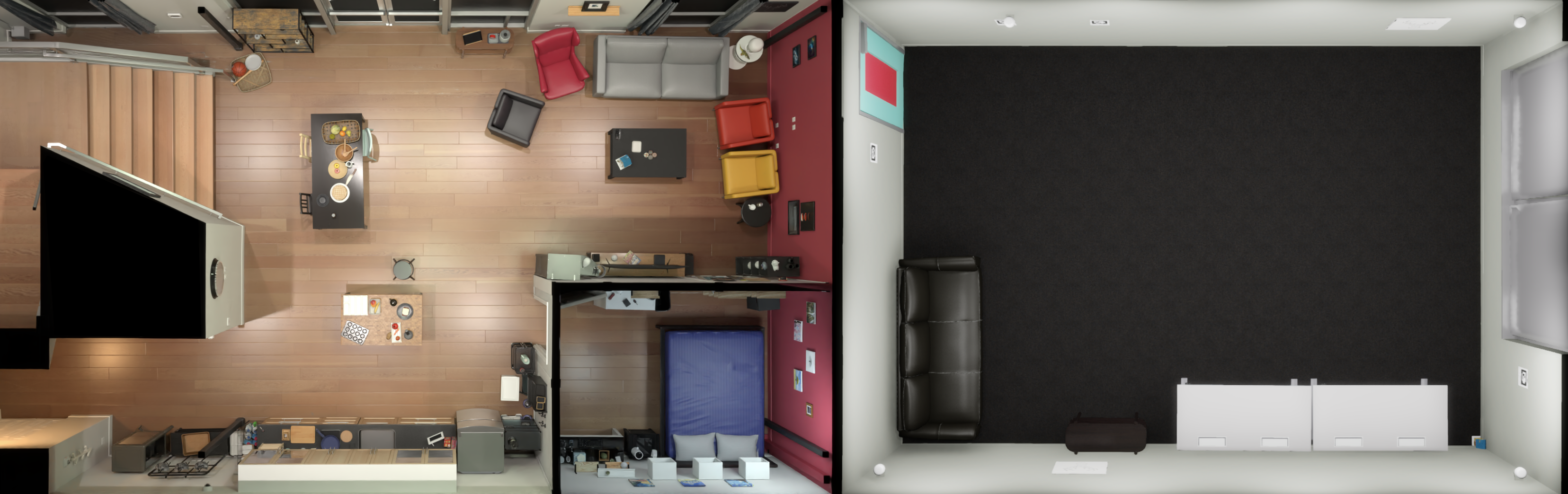}
    \caption{Top-down rendering of two spaces with the apartment on the left and the office on the right.}
    \label{fig:dataset_overview_1-top}
  \end{subfigure}
  \hfill
  \begin{subfigure}[b]{\textwidth}
    \includegraphics[width=\textwidth]{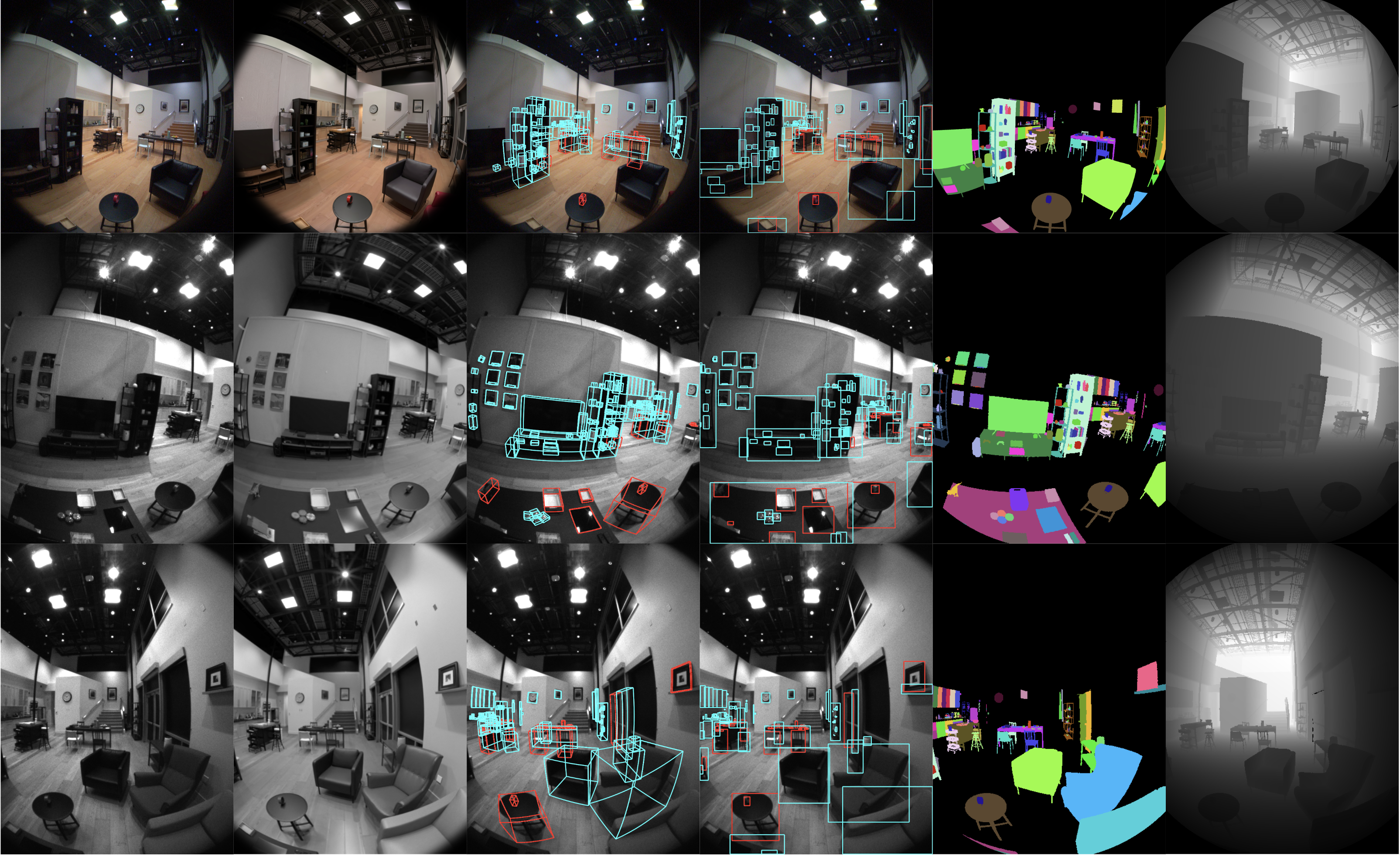}
    \caption{2D visualization of ground truth projected onto Aria camera sensors. From top to bottom: the RGB, the left monochrome, the right monochrome camera sensors. From left to right: raw sensor image; photo-realistic synthetic rendering; 3D bounding boxes (cyan for stationary and red for dynamic objects), 2D bounding boxes, segmentation masks for all object instances; depth map.}
    \label{fig:dataset_overview_1-bottom}
  \end{subfigure}
\end{center}
  \caption{An overview of the ADT dataset.}
\label{fig:dataset_overview_1}
\end{figure*}

\begin{abstract}
 We introduce the Aria Digital Twin (ADT) - an egocentric dataset captured using Aria glasses with extensive object, environment, and human level ground truth. This ADT release contains 200 sequences of real-world activities conducted by Aria wearers in two real indoor scenes with 398 object instances (324 stationary and 74 dynamic). Each sequence consists of: a) raw data of two monochrome camera streams, one RGB camera stream, two IMU streams; b) complete sensor calibration; c) ground truth data including continuous 6-degree-of-freedom (6DoF) poses of the Aria devices, object 6DoF poses, 3D eye gaze vectors, 3D human poses, 2D image segmentations, image depth maps; and d) photo-realistic synthetic renderings. To the best of our knowledge, there is no existing egocentric dataset with a level of accuracy, photo-realism and comprehensiveness comparable to ADT. By contributing ADT to the research community, our mission is to set a new standard for evaluation in the egocentric machine perception domain, which includes very challenging research problems such as 3D object detection and tracking, scene reconstruction and understanding, sim-to-real learning, human pose prediction - while also inspiring new machine perception tasks for augmented reality (AR) applications. To kick start exploration of the ADT research use cases, we evaluated several existing state-of-the-art methods for object detection, segmentation and image translation tasks that demonstrate the usefulness of ADT as a benchmarking dataset.
\end{abstract}

\section{Introduction}
Egocentric data has become increasingly important to the machine perception community in the past several years due to the rapid emergence of AR applications. Such applications require the co-existence of the real-world space and a virtual space along with a contextual awareness of the real surroundings. Complete contextual awareness cannot be achieved without a full and accurate 3D digitization of three fundamental elements in the real-world space: humans, objects and the environment. Every object and environmental component, including lighting, room structure and layout, has to be precisely digitized to unlock consistent rendering of the virtual space within the real world. Dynamic object motion needs to be tracked in 3D to update the state of the space via physical interactions. The state of the human wearing AR glasses should be estimated and intersected with the digital space to derive the interaction in both physical and virtual spaces. Achieving all of this requires solutions to a number of core problems such as 3D object detection, human pose estimation, and scene reconstruction, where \emph{data} is the key component. 

Existing datasets that aim at progressing the field of AR do not focus holistically on the problem space, but rather on specific sub-problems. A significant amount of progress in large scale static scene datasets~\cite{dai2017scannet,sun-rgbd,Matterport3D} has helped to advance 3D scene understanding tasks such as static object detection, scene reconstruction and room layout estimation. Although the photo-realism of these reconstructed scenes is continuously improving~\cite{replica19arxiv}, these datasets lack the motion of objects introduced by hand interactions that commonly occur in egocentric AR scenarios. Object-centric datasets~\cite{objectron2021,xiang2017posecnn} that include increasingly complex occlusions between objects, also require that objects be stationary to facilitate the annotation process. Dynamic object datasets~\cite{FirstPersonAction_CVPR2018,garon2018framework,hampali2020honnotate} capture hand-object interaction but the data is captured in controlled, simplified environments. Egocentric human motion datasets~\cite{rhodin2016egocap,zhao2021egoglass} capture 3D human poses with annotation of 3D joint positions but without the digitization of the environment. Most importantly, none of the discussed datasets leverage an AR-style sensing device that captures the unique challenges with egocentric data such as fast ego motion, sub-optimal viewpoint, low-power sensing hardware, etc. Although some egocentric datasets~\cite{grauman2022ego4d,tang2023egotracks,TREK150ijcv} have emerged recently, they present only either narrative annotation or 2D object annotation without addressing the challenges in 3D space.

The availability of egocentric data capture devices has been surging in recent years, e.g., Vuzix Blade, Pupil Labs, ORDRO EP6, etc. Among them, the popularity of Aria glasses is quickly growing due to its standard glasses-like form factor and the full egocentric sensor suite including, but not limited to, a red-green-blue (RGB) camera, two monochrome cameras, two eye tracking cameras and two inertial measurement units (IMUs) which allows users to tackle a broad spectrum of machine perception tasks in real-world activities. The availability of Aria data has been accelerated by the recent release of the Aria Pilot Dataset~\cite{aria_pilot_dataset}. 

\begin{figure}[!t]
\begin{center}
  \includegraphics[width=\linewidth]{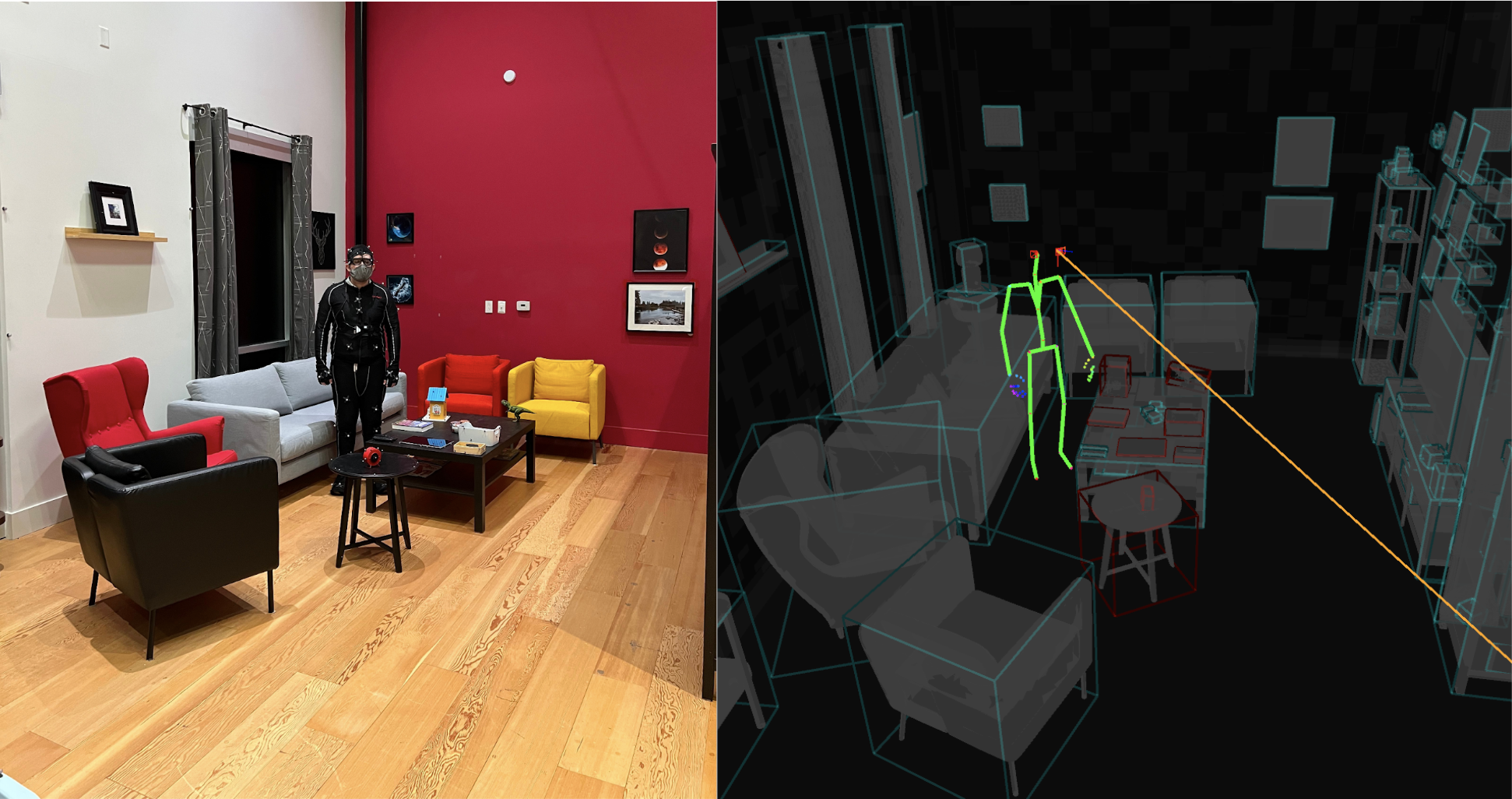}
\end{center}
\caption{Left: A snapshot of the data recording process in the apartment. Right: A 3D visualization of ground truth for object bounding boxes and the collector's body skeleton in green with eye gaze shown in orange.}
\label{fig:dataset_overview_2}
\end{figure}

Motivated by the gap in holistic egocentric 3D data highlighted above, we have created the Aria Digital Twin (ADT) dataset to accelerate egocentric machine perception research for AR applications. This dataset offers 200 sequences collected by Aria-wearers performing real-world activities in two realistic spaces - an apartment and an office, with a combined total of 350 stationary and 50 dynamic object instances. Compared to existing work, each ADT sequence offers more complete and accurate ground truth data for the digital space including: device calibrations, device and object 6-degree-of-freedom (6DoF) poses, human poses, eye gaze vectors, object segmentation, depth maps and photo-realistic synthetic images. Figure~\ref{fig:dataset_overview_1-top} shows top-down renderings of two spaces, and Figure~\ref{fig:dataset_overview_1-bottom} shows a 2D visualization of all object ground truth projected onto the Aria camera sensors. Figure~\ref{fig:dataset_overview_2} shows a snapshot image of the data capturing process and a 3D rendering of the human ground truth data. 

To build this dataset, we reconstructed every object and the entire environment of the two spaces in a metric, photo-realistic pipeline. We integrated a motion capture system with the digitized space and precisely synchronized it with the Aria glasses to track objects and humans while recording egocentric data in a spatio-temporally aligned environment. We demonstrate the quality of the 3D reconstruction via a qualitative evaluation and the accuracy of the object tracking via a novel quantitative evaluation. We performed evaluations on several existing state-of-the-art methods for object detection, segmentation and image translation tasks to demonstrate the usefulness of ADT when testing AR related machine perception algorithms. Our contribution is the establishment of a new standard for both the quality and comprehensiveness of digitized real-world indoor spaces to advance fundamental AR research by means of an exemplary dataset and methodology for the creation of such a dataset. 

\section{Related Work}
\label{section:related_work}
Several works have been published that only provide static object 3D poses. Objectron~\cite{objectron2021} contains object-centric short videos captured by mobile phones with 6DoF pose annotations over nine categories of objects for 3D object detection tasks. The objects remains stationary and rapid movement of the camera is avoided. The BOP challenge~\cite{hodan2018bop} is composed of several datasets for 6DoF object pose estimation. The LM~\cite{hinterstoisser2013model} and LM-O~\cite{brachmann2014learning} datasets provide 6DoF poses of objects in stationary scenes in the form of 15 cluttered objects placed on a table top with markers. T-LESS~\cite{hodan2017t} introduces the challenge of handling textureless objects in the 3D pose estimation problem. YCB-V~\cite{xiang2017posecnn} provides RGB-D videos of 21 stationary cluttered objects by means of a semi-automatic annotation process. All of the above works ignore human interactions with objects which limits their utility for many real-world AR problems.

Another common type of dataset focuses on only dynamic object 3D poses. FPHA~\cite{FirstPersonAction_CVPR2018} provides both hand pose and 6DoF object pose annotation for 25 dynamic moving objects captured by a shoulder-mounted RGB-D sensor. The object poses are coarsely estimated by a magnetic sensor placed close to the approximated center of mass. ECVA~\cite{garon2018framework} and Ho-3D~\cite{hampali2020honnotate} offer 6DoF object pose annotation for dynamic objects in a table mounted RGB-D sensing setup. TUD-L~\cite{hodan2018bop} contains only 3 dynamic objects with semi-automatically annotated 6DoF pose. Although hand-object interaction and occlusion are provided, these datasets have a limited number of objects, and do not have head-mount capture devices, limiting their relevance to AR-based machine perception. HOI4D~\cite{liu2022hoi4d} and H2O~\cite{Kwon_2021_ICCV} presented data captured by a head-mounted RGB-D sensor, annotated with 2D segmentation and 3D object pose. However, the level of geometric accuracy for the scene and objects, which is reconstructed using low fidelity sensors, falls short of that provided in ADT. Additionally, the absence of static scene modeling and photo-realistic reconstruction prevents them from being used for many AR tasks that need to bridge the real and virtual world gap. 

Many works summarized above completely omit discussion of how accurate their ground truth data may or may not be, and those that have made an attempt, usually use subjective approaches, or approaches prone to human error and variability. For example, Objectron~\cite{objectron2021} compares annotations across different annotators to see how much variability there is in the results. They also only present results run on a small set of the object types which means other objects may have very different variability in ground truth data. Ho-3D~\cite{hampali2020honnotate} does a similar validation procedure where they manually annotate point clouds from the RGB-D sensors, but this is also prone to human subjectivity error as well as sensor error since RGB-D sensors have errors on the order of centimeters. 

A number of egocentric video datasets have been recently released that capture realistic activities. EpicKitchen~\cite{Damen2018EPICKITCHENS,Damen2022RESCALING} and Charades-Ego~\cite{Sigurdsson2018ActorAO} record a wearer's daily indoor activities and annotate the data with action segments and 2D object bounding boxes. EGTEA~\cite{li2021eye} contains eye gaze attention in addition to activity annotation. Ego4D~\cite{grauman2022ego4d} builds a notably large dataset partially composed of audio, 3D meshes of the environment and eye gaze with multi-camera sensors. TREK-150~\cite{TREK150ijcv} and EgoTracks~\cite{tang2023egotracks}, focusing on hand-object interactions, are composed of egocentric videos with objects annotated by their 2D bounding boxes. Annotation in each of the above mentioned datasets are at 2D image level without any understanding of the 3D world. Other datasets such as EgoCap\cite{rhodin2016egocap} and EgoGlass\cite{zhao2021egoglass} are proposed to address the egocentric human pose estimation task. However, the complexity of the environment and interactions with objects are ignored. Mo2Cap2~\cite{xu2019mo} and UnrealEgo~\cite{hakada2022unrealego} introduce environment complexity but are generated synthetically.

Scene datasets such as SUN-RGB-D~\cite{sun-rgbd}, ScanNet~\cite{dai2017scannet} and Matterport3D~\cite{Matterport3D}, provide reconstructions of large scale real indoor scenes. Videos of these scenes are typically recorded using RGB-D cameras. The videos are annotated with 2D segmentation, 3D object bounding boxes and semantic scene information. Although they provide scene level ground truth, all objects in the scenes are static and the capturing device is not egocentric. Their scene digitization is also not optimized against reality and hence does not meet the photo-realism bar. This limits the utility of these datasets in training systems for the real world. Replica~\cite{replica19arxiv} significantly improves on the reconstruction quality aspect but is once again not egocentric. Synthetic scene datasets such as HyperSim~\cite{hypersim} and Openrooms~\cite{li2020openrooms} gather high-quality 3D models online and fine-tune the models in post-processing to create visually convincing scenes. They do not have a real-world counterpart recordings so the gap between simulated and real data remains. Furthermore, the lack of egocentric data in these spaces does not allow researchers to use these for solving AR tasks.
 
\section{Dataset Generation Methodology}
\label{section:dataset_gen}

Our dataset generation procedure starts by creating a stationary, photo-realistic digital scene followed by enabling the tracking of Aria glasses, objects and humans within the scene. 

\subsection{Stationary Scene Digitization}
\label{section:dataset_gen:scene}

\begin{figure}[!tbp]
\begin{center}
  \begin{subfigure}[b]{0.23\textwidth}
    \includegraphics[width=\textwidth]{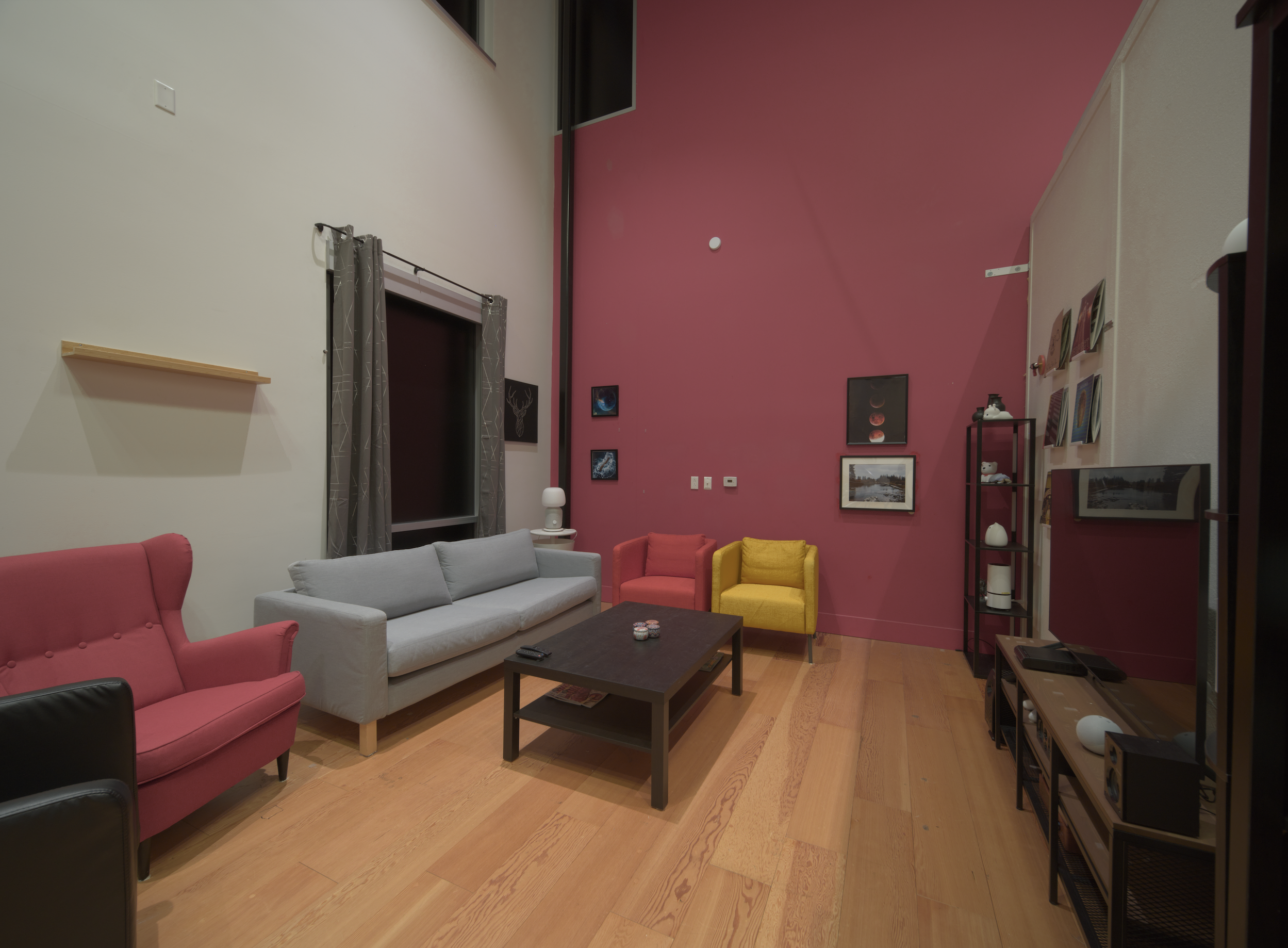}
    \caption{Real photo of a room.}
    \label{fig:photorealism_room_real}
  \end{subfigure}
  \hfill
  \begin{subfigure}[b]{0.23\textwidth}
    \includegraphics[width=\textwidth]{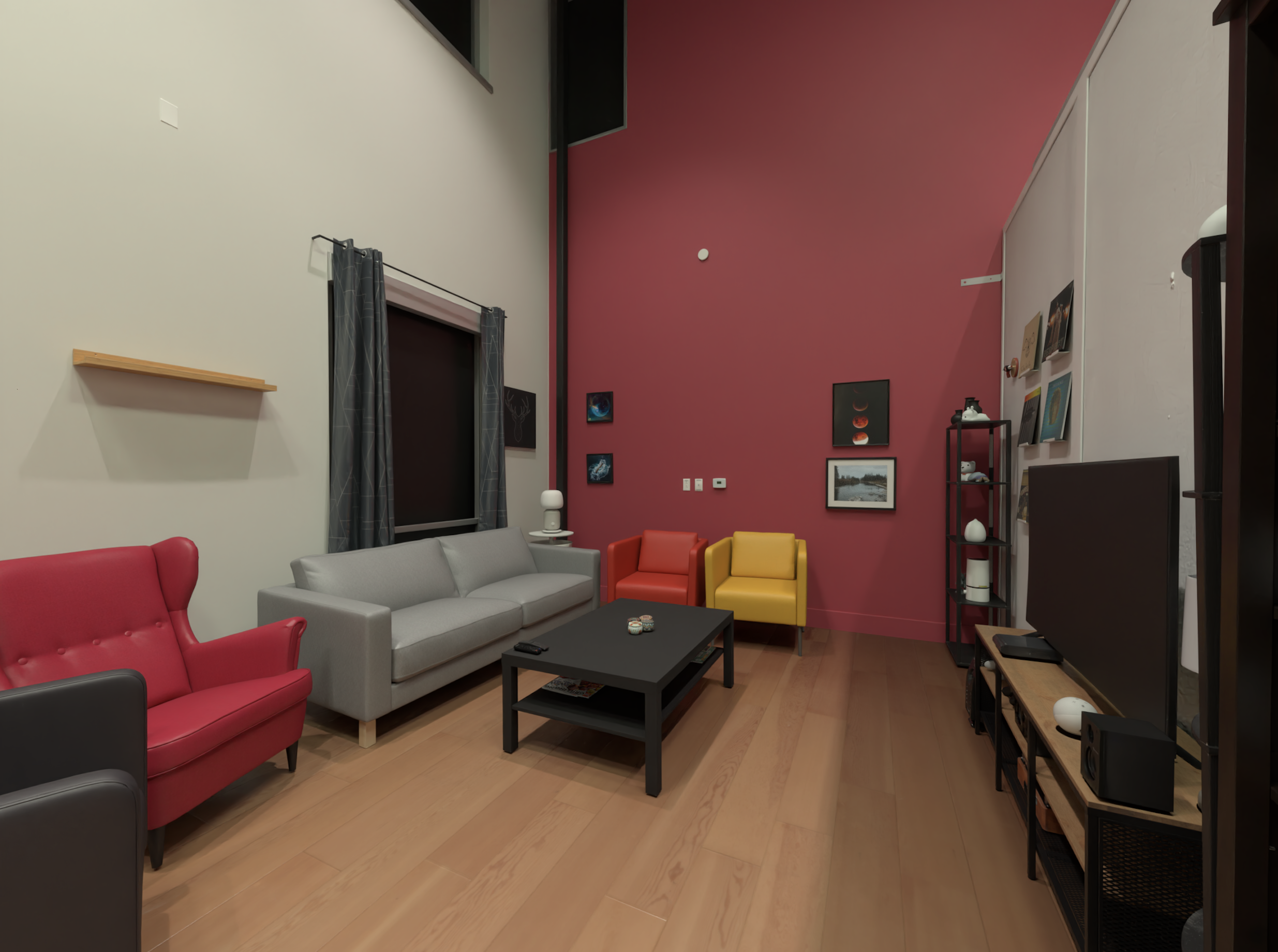}
    \caption{Rendering of a room.}
    \label{fig:photorealism_room_render}
  \end{subfigure}
  \begin{subfigure}[b]{0.23\textwidth}
    \includegraphics[width=\textwidth]{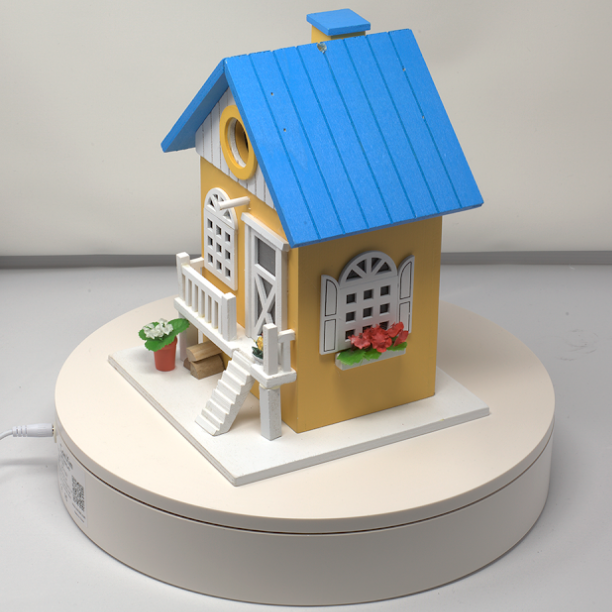}
    \caption{Real photo of a birdhouse.}
    \label{fig:photorealism_object_real}
  \end{subfigure}
  \hfill
  \begin{subfigure}[b]{0.23\textwidth}
    \includegraphics[width=\textwidth]{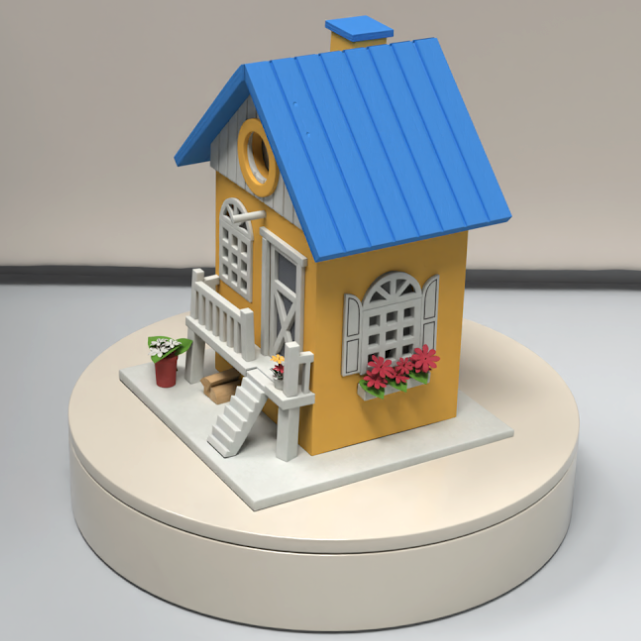}
    \caption{Rendering of a birdhouse.}
    \label{fig:photorealism_object_render}
  \end{subfigure}
  \caption{Real photos and their synthetic counterparts used to optimize the empty room digitization and individual object digitizations.}
  \label{fig:real_vs_syn}
\end{center}
\end{figure}

\textbf{Room digitization}: Taking the apartment as an example, the physical space is first emptied and scanned using a high-resolution scanner - FARO Focus S-150. The generated point cloud is then converted to a triangular mesh by fitting planes based on the room topology. The error of the meshing process is measured against the raw source point cloud using a closest-point-to-mesh distance metric, resulting in a total 50th and 80th percentile (P50 \& P80) error of 0.688mm and 4.68mm respectively. We also reconstruct Physically-Based Rendering (PBR) materials including albedo, roughness and metallic maps. Albedo maps are extracted via photogrammetric reconstruction. Roughness and metallic values are manually assigned to different portions of the space based on material properties such as metal, glasses, etc. Each light source in the scene is parameterized by intensity, shape and color, and is tuned manually by taking diffuse and chrome spheres as references. To make sure the reconstructed materials and lighting are accurate enough to deliver photo-realistic quality, we implement a fully digital rendering using Nvidia's Omniverse path tracing software and iteratively tune all of the material and lighting parameters against a real photographic reference frame as shown in Figure ~\ref{fig:photorealism_room_real} and ~\ref{fig:photorealism_room_render}.

\textbf{Object Digitization}: The geometry of each object is acquired using the ATOS 5 Bluelight 3D scanner, which provides geometry data to an industrial standard for manufacturing. The material is reconstructed through a photogrammetry process, similar to the room digitization process, but in a photo booth setup consisting of a turn table, four LED panels and three Canon 5D Mark IV cameras with cross-polarization used to eliminate specularity of the material. Also similar to the room digitization process, we setup a real-vs-synthetic comparison to tune the material of the object to match the real photo as shown in Figure ~\ref{fig:photorealism_object_real} and ~\ref{fig:photorealism_object_render}. 

\textbf{Layout Digitization}: After gathering the 3D models for the room and objects, we physically furnished the space and set large furniture pieces to be stationary objects as they are not typically moved in day-to-day real-world scenarios. We then perform a new FARO scan of the fully furnished space, initialize the 6DoF objects poses by manually placing the 3D models into the point cloud and use Iterative Closest Point (ICP)~\cite{segal2009generalized} to optimize the geometry alignment. Similar to the room digitization process, we monitored the quality using the closest-point-to-mesh distance metric and achieved 4.67mm at P50 and 20.51mm at P80 representing the combined geometrical error from object digital models and the layout for the entire scene.

\subsection{Pose Generation}
\label{section:dataset_gen:pose}

\begin{figure}[t]
\begin{center}
  \includegraphics[width=0.8\linewidth]{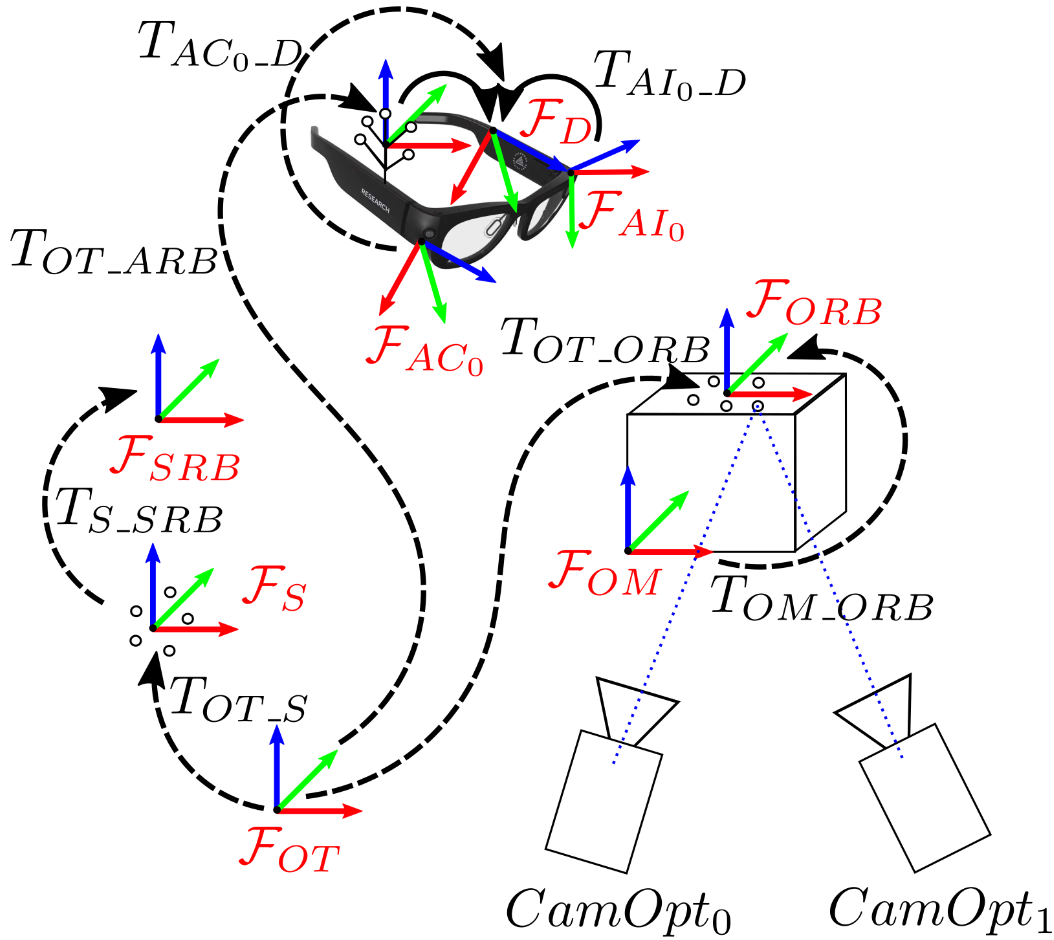}
\end{center}
\caption{ADT Scene System Diagram.}
\label{fig:sys_diag}
\end{figure}

We track 3D poses of three dynamic components in an ADT space: objects, Aria glasses and human (Aria wearers). All of them are expressed in a single Scene frame of reference for all sequences captured in the same ADT space, namely, $F_{S}$. This allows us to plot object, device and human poses from multiple captures collected at different time frames or across different devices, in the same coordinate space. For simplicity, we make $F_{S}$ the same frame of reference as the one used in the stationary scene digitization process explained above so the 3D poses for stationary objects are determined without an additional conversion. Figure~\ref{fig:sys_diag} illustrates an example configuration with one dynamic object (drawn as a cube), one Aria device and two example Optitrack cameras. Figure~\ref{fig:sys_diag} also shows all the relevant frames of reference in a single ADT space, as well as the system measurements used to provide the final pose estimates. Note that since the final data contains all poses relative to the Scene frame, all Optitrack frames are removed from the final data. Our pose generation process relies on the Optitrack motion capture system, which provides high rate sub-millimeter level precision poses~\cite{aurand2017accuracy}, to track dynamic object and Aria poses. 

\textbf{Dynamic Object Pose}: Each object $k$, tracked by Optitrack, has its own coordinate frame that defines the rigid body (RB) of that object. A RB is created by a set of markers rigidly attached to the object and its 3D pose in the Optitrack's frame, $F_{OT}$, is represented as $T_{OT\_ORB_k}$\footnote{$T_{B\_A}$ is a special Euclidean group (SE(3)) transformation matrix that transforms coordinate frame A to coordinate frame B, expressed in frame B.}. For an object $k$, our goal is to calculate $T_{S\_OM_{k}}$ expressed in Eqn.~\ref{eqn:dyn_obj_pose}, where $F_{OM}$ is the object model's frame set during scene digitization. To calculate the pose between each dynamic object's RB frame and its model frame ($T_{ORB_{k}\_OM_{k}}$), we scan each object twice, one with markers installed and the other without. We then register two generated meshes using point-set registration. To convert coordinates in $F_OT$ to $F_S$, we create a scene RB for each ADT space by installing markers on the walls, followed by computation of $T_{S\_OT}$ by aligning the scan-extracted 3D marker positions to the Optitrack measured scene RB points using a point-set registration method similar to ICP~\cite{segal2009generalized}. 

\begin{equation}
\begin{split}
    T_{S\_OM_k} = &  T_{S\_OT} \times  T_{OT\_ORB_k} \times T_{ORB_k\_OM_k}\\ 
\end{split}
\label{eqn:dyn_obj_pose}
\end{equation}

\textbf{Aria Device Pose}: Similar to the dynamic object pose generation process, we use Optitrack to track each Aria device's RB frame, $F_{ARB}$, relative to $F_{OT}$, and then compute the pose of Aria's device frame, $F_D$, relative to the $F_S$ ($T_S\_ARB$). We start with estimating the SE(3) transform from one IMU frame, $F_{AI_0}$, to $F_{ARB}$ ($T_{ARB\_AI_0}$). This is estimated by collecting a dataset where we excited the device about all 6DoF for approximately one minute while Optitrack is tracking the RB. We fit the IMU data to a trajectory, and solve for the $T_{ARB\_AI_0}$ that best aligns this IMU trajectory to Optitrack's measured Aria RB trajectory. We further calibrate each Aria's extrinsics and intrinsics including: 1) SE(3) transforms between all sensor frames and the device frame, 2) calibrated camera models using Kannala Brandt~\cite{usenko2018double} and fisheye radial-tangental thin prism~\cite{weng1992camera} parameterizations, and 3) calibrated linear rectification models for both accelerometers and gyroscopes.

\begin{equation}
\begin{split}
    T_{S\_AI_{0}} = &  T_{S\_OT} \times  T_{OT\_ARB} \times T_{ARB\_AI_{0}}\\ 
\end{split}
\label{eqn:aria_pose}
\end{equation}

Equally important to device calibration is Optitrack-Aria time synchronization. We employ a continuous synchronization strategy based on the Society of Motion Picture and Television Engineer's SMPTE timecode, a widely used standard for synchronized timing between audio and video captures in the motion pictures industry. Our timecode solution uses a set of UltraSync One devices made by Timecode Systems which synchronizes our Optitrack machine to all Aria devices, achieving a measured average accuracy of less than 10 microseconds according to our own Aria-Optitrack specific tests.

\textbf{Human Pose}: To track a person during data collection, we use the Biomechanic57 template provided by Optitrack's Motive software which estimates the human skeleton using a set of markers placed at specific locations on the body. We output the human joints estimated by Motive, as well as the raw marker positions for researchers to perform their own body pose estimation. We also use the raw marker positions to compute 3D body meshes using our proprietary software.

\subsection{System Accuracy}
\label{sec:sys_acc}

We propose a novel evaluation pipeline for measuring the total system error in our 3D object pose ground truth data generation system. We argue that this proof of fidelity significantly improves the value of our dataset as it gives researchers, for the first time, additional signal as to the expected performance of their algorithms built off our data.

\textbf{Methodology}: Since object ground truth data in ADT is correlated to Aria images, we propose to quantify the object pose error, $e_p$, and the reprojection error, $e_r$ of objects within view of any of the Aria images. Eqn.~\ref{eqn:error-rep-def} describes the reprojection error of an $i^{th}$ point from object $k$ projected into the image plane of camera $j$ using the object pose and calibrated camera model. We denote $\hat{T}$ as the measured object pose, $T$ as the true object pose, $\pi$ as the camera projection model which maps $\mathbb{R}^3 \xrightarrow[]{} \mathbb{R}^2$, $\kappa$ as the calibrated intrinsic parameters, and finally $P_{OM_k}^i$ is the position of marker $i$ expressed in the model frame of object $k$. Eqn.~\ref{eqn:error-pose-def} describes the pose error for each $k^{th}$ object in each camera $j$. Since $e_p$ in Eqn.~\ref{eqn:error-pose-def} is an SE(3) transformation between the true and measured frames, we extract the translation and rotation errors as scalar values using the L2 norm of the translation and the magnitude of the angle value from an Axis-Angle rotation representation. 

\begin{equation}
    e_{r_{k, j, i}} = \pi(\hat{T}_{C_j\_OM_k} \times P_{OM_k}^i, \kappa) - \pi(\hat{T}_{C_j\_OM_k} \times p_i, \kappa)
\label{eqn:error-rep-def}
\end{equation}

\begin{equation}
    e_{p_{k, j}} = \hat{T}_{C_j\_OM_k} \times {[T_{C_j\_OM_k}]}^{-1}
\label{eqn:error-pose-def}
\end{equation}

In Eqns.~\ref{eqn:error-rep-def} \&~\ref{eqn:error-pose-def}, $\hat{T}_{C_j\_OM_k}$ is known from Eqns.~\ref{eqn:dyn_obj_pose} \&~\ref{eqn:aria_pose}, therefore the only unknowns are the true object poses ${T}_{C_j\_OM_k}$. To compute the true object poses relative to Aria images, we propose labeling Optitrack markers in Aria images, and finding the object pose that minimises reprojection error between estimated marker projections and labeled pixels. Since we have precise measurements of the object pose from the ground truth results, we can create a non-linear optimizer initialized with values close to the true values to ensure a high likelihood of convergence to a global minimum. We therefore define our objective function, $\Phi$, as shown in Eqn.~\ref{eqn:error-obj}, where $U_{d_i}$ is a $2 \times 1$ vector of labeled marker pixels. We minimize $\Phi$ to solve for the object pose using a Levenberg-Marquardt optimizer with a Huber Loss function. Eqn.~\ref{eqn:error-obj} shows the objective function for object $k$, camera $j$, and at a specific Aria frame time.

\begin{equation}
    \Phi_{C_j, OM_k}(T_{C_j\_OM_k}) = \sum_{i=1}^{I} [ U_{d_i} - \pi(T_{C_j\_OM_k} \times P_{OM_k}^i, \kappa)]
\label{eqn:error-obj}
\end{equation}

\textbf{Results}: The validation dataset consists of a recording for each dynamic object used in ADT. To minimize sources of error due to marker labeling in our proposed validation methodology, we hold the objects within arms length of Aria during data capture. Since all pose data is captured from Optitrack cameras, the system error is independent of the object distance away from the camera, therefore collecting validation data of far away objects would provide no additional benefit. We collect such validation sequences in both ADT spaces, with multiple Aria devices, where the Aria and the objects are both moving at similar rates as would be expected in the regular dataset releases. We then run approximately 10 frames of each object through our validation pipeline. Table~\ref{table:sys_acc_results} shows a summary of the final system accuracy results. The results show average errors of 6.78 pixels (measured with Aria RGB images at 1408x1408 resolution, 110 deg field of view), 1.29 deg and 6.83 mm for the measured reprojection error, rotation error, and translation error, respectively. It is important to note that measured reprojection error is larger  than should be expected for regular datasets since we only extract measurements when the object is close to the Aria camera, resulting in a higher than average reprojection error in pixel units. We also include the resulting optimized reprojection error, which is the reprojection error after optimizing for the real object pose to prove that our methodology generates accurate real poses.

\begin{table}
\begin{tabular}{|c|c|c|c|c|}
\hline
      & Measured      & Optimized     & Rot    & Trans \\
      & Proj[pixel] & Proj[pixel] & [deg]  & [mm] \\
\hline
Average & 6.78 & 0.56 & 1.29 & 6.83 \\ \hline
Median & 6.00 & 0.46 & 0.91 & 5.18 \\ \hline
\end{tabular}
\caption{System accuracy results for all dynamic objects.}
\label{table:sys_acc_results}
\end{table}
        
\subsection{Data Annotations}
\label{section:dataset_gen:annotations}
In this section, we will describe how the remaining ground truth data is derived from the raw Aria sensor data and digital scene models along with the poses of 3D objects, Aria glasses and human bodies.

\begin{figure}[!tbp]
\begin{center}
  \includegraphics[width=0.8\linewidth]{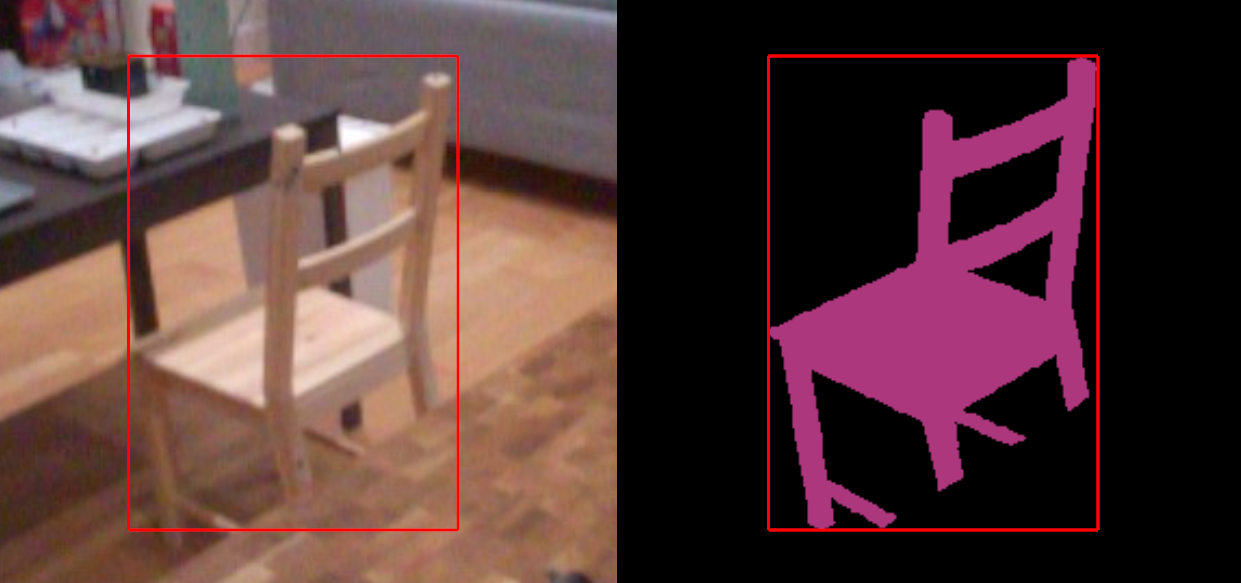}
  \caption{Occlusion between the chair and the table is accounted for in the ground truth segmentation and 2D bounding box.}
  \label{fig:render:occlusion}
\end{center}
\end{figure}

As described in Section~\ref{section:dataset_gen:pose}, for every frame captured by the Aria device we have the poses of all objects as well as the cameras and wearer within the scene. Coupled with the calibration parameters, this completes a full generative model that can be used to render a fully synthetic equivalent for every captured frame, as shown in Figure~\ref{fig:real_vs_syn}. We leverage a custom shader\footnote{A shader is a small program that runs per-pixel during a typical graphics rasterization routine.} that instead of rendering object texture, renders the unique object integer IDs and metric depth per-pixel, for per-frame instance-level segmentation and depth respectively. We then directly calculate the 2D axis-aligned bounding boxes of each object instance in each image based on the segmentation image from the above process. This process results in ground truth 2D segmentations, depth maps, and 2D bounding boxes for each image frame.  The dynamic object-to-object occlusion is automatically taken care of in this process. Figure~\ref{fig:render:occlusion} shows an example of such cases. We also apply the same process to human-to-object occlusion cases using the approximated body mesh. 

Furthermore, we provide eye gaze estimates using Aria eye tracking camera images collected at a rate of 30Hz. Each pair of eye tracking images is processed using our proprietary eye tracking software to produce a per frame gaze direction vector. We then compute the ray depth by finding the intersection with the scene objects.

\section{Dataset Content}
\label{section:dataset_content}
The ADT dataset was recorded in two spaces: an apartment and an office environment. The apartment is composed of a living room, kitchen, dining room and bedroom, whereas the office space is a single room with very minimal office furniture. The apartment has 281 unique stationary objects and the office room has 15 unique stationary objects. Given some objects have multiple instances that may differ slightly, the apartment has a total of 324 stationary object instances and the office room has 20 stationary object instances. In addition, there are 74 single-instance dynamic objects shared between two spaces.   

\begin{figure}[t]
\begin{center}
  \includegraphics[width=0.8\linewidth]{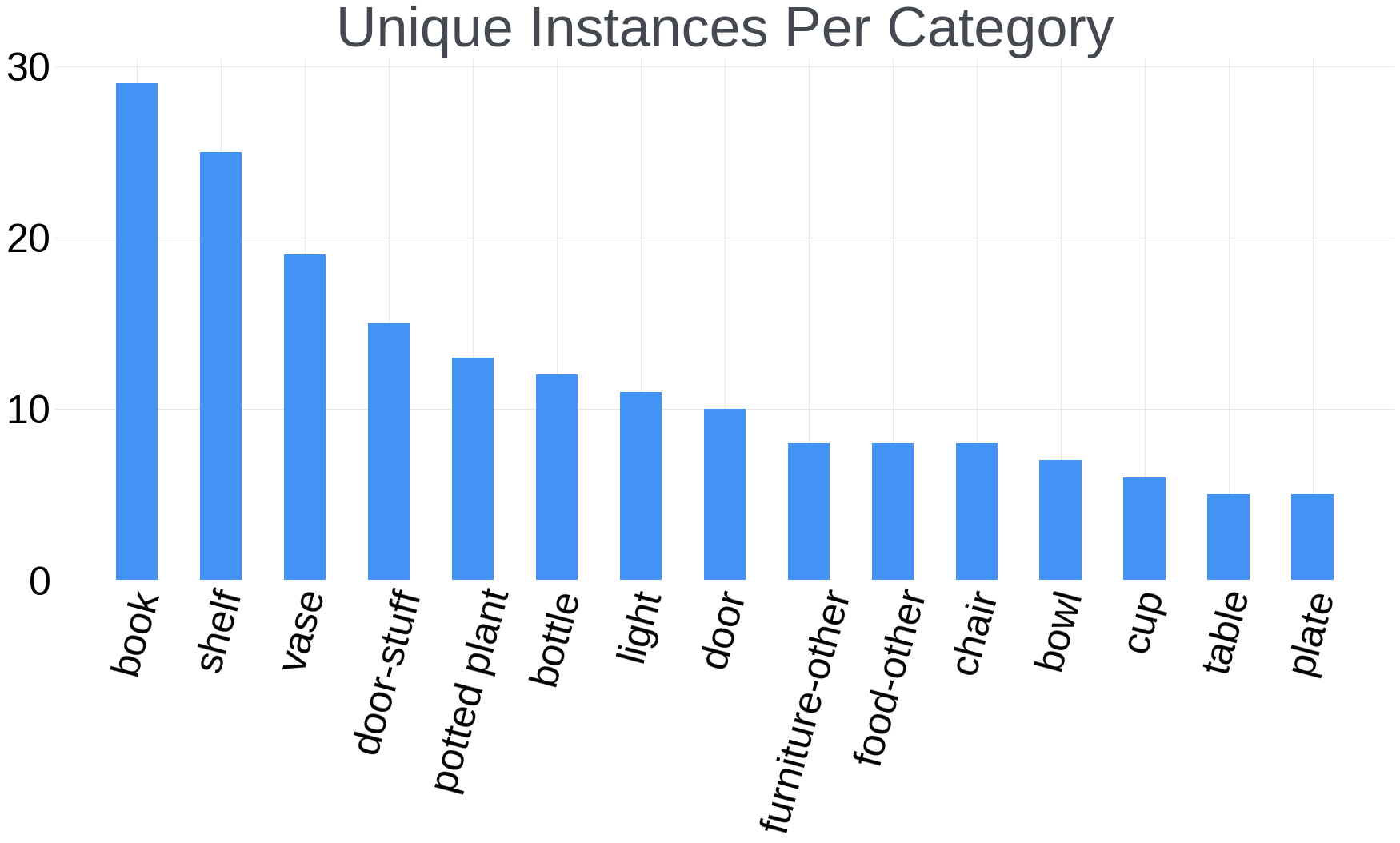}
\end{center}
\caption{Number of unique object instances for the top 15 categories following the COCO definition.}
\label{fig:object_category_distribution}
\end{figure}

Strong emphasis was put on the realness of the ADT spaces and the diversity of objects so that we could collect data in plausible real-life scenarios instead of contrived laboratory situations. We generated a list of common activities in these two spaces under the envisaged setting and selected appropriate objects for these activities. Each object is annotated with its category. The histogram of the top 15 object categories is shown in Figure~\ref{fig:object_category_distribution} following the category definition from the COCO~\cite{lin2014microsoft} dataset. 


We release 200 sequences in total with 150 sequences in the apartment and 50 sequences in the office room. We designed 5 single-person activities and 3 dual-person activities in the apartment. The single-person activities are room decoration, meal preparation, work, object examination and room cleaning. The dual-person activities include partying, room cleaning and dining table cleaning. Every activity has 10 to 50 sequences which captures an abundance of variation in the collectors' motion and object interactions. For the office dataset, we include object examination as the single-person activity. 

\begin{table}[!t]
\begin{center}
\begin{tabular}{|c|c|c|}
\hline
 & AP-box & AP-Mask  \\ \hline
FPN & 21.36  &  19.81  \\ \hline
VIT-B & 11.42  & 11.64   \\ \hline
\end{tabular}
\end{center}
\caption{AP-box and AP-Mask (in \%) for the 2D object detection and image segmentation tasks on the ADT dataset.}
\label{table:benchmarking_2d}
\end{table}

\section{Benchmarking}
\label{section:benchmarking}
Having created a richly annotated dataset, we perform an evaluation of various state-of-the-art methods for AR related tasks including 2D object detection, 2D image segmentation, 3D object detection and image to image translation. With these experiments, we show that our dataset is well suited for evaluating important perception tasks while also aiming at inspiring new machine perception use cases.  

\subsection{2D Object Detection and Image Segmentation}
\label{sec::2d_tasks}
We select two state-of-the-art methods based on their performance on the MS-COCO ~\cite{lin2014microsoft} and LVIS ~\cite{gupta2019lvis} datasets: the Feature Pyramid Network (FPN) ~\cite{lin2017fpn}, a seminal work using hierarchical backbones; and the VIT-Det~\cite{li2022vit}, a transformer-based non-hierarchical backbone framework. Both methods are tested on rectified Aria RGB images to maintain the consistency with their models pre-trained on MS-COCO. To perform the evaluation, we map ADT objects into relevant categories in the MS-COCO taxonomy. We adopt the box average precision (AP-Box) and mask average precision (AP-Mask) defined in COCO evaluation protocol ~\cite{lin2014microsoft}. We aggregate the results for all frames in each sequence and then average them across all ADT sequences. The evaluation results, shown in Table ~\ref{table:benchmarking_2d}, highlight the domain gap between models trained on MS-COCO, a popular large-scale training dataset, and real world egocentric data present in the ADT. This poor performance may be attributed to the fast ego motion and sub-optimal viewpoint in the ADT data, which was also observed by TREK-150 ~\cite{TREK150ijcv} for the 2D object tracking task on egocentric videos. 

\subsection{3D Object Detection}
We evaluate two state-of-the-art 3D object detection methods, Total3D~\cite{Nie_2020_CVPR} and Cube R-CNN~\cite{brazil2022omni3d}, pre-trained on ScanNet~\cite{dai2017scannet} and Omni3D~\cite{brazil2022omni3d}, respectively. Both methods are tested on rectified Aria RGB images similar to the tasks in Section ~\ref{sec::2d_tasks}. Since Total3D requires 2D bounding box input, we select MaskRCNN as its 2D detector for a fair comparison. Similar to Omni3D, we adopt average precision (AP) as the metric. We compute the AP across all sequences covering 7 object categories\footnote{The common categories between COCO2017, NYU and Omni3D are television, book, refrigerator, sofa, bed, chair, table.} and 1.6 million GT 3D bounding boxes in total, with a confidence threshold of 0.2 and IoU threshold at 0.25. The AP numbers of the top five categories are reported in Table~\ref{table:benchmarking_3d}. The results indicate the similar challenges to the tasks in Section ~\ref{sec::2d_tasks}. Additionally, we observe that the monocular sensor input, required by both Total3D and Cube R-CNN, often yield wrong depth for object 3D poses that can be potentially improved by using Aria's multi-camera sensors.

\subsection{Image to Image Translation}
\begin{figure}[t]
\begin{center}
  \includegraphics[width=0.9\linewidth]{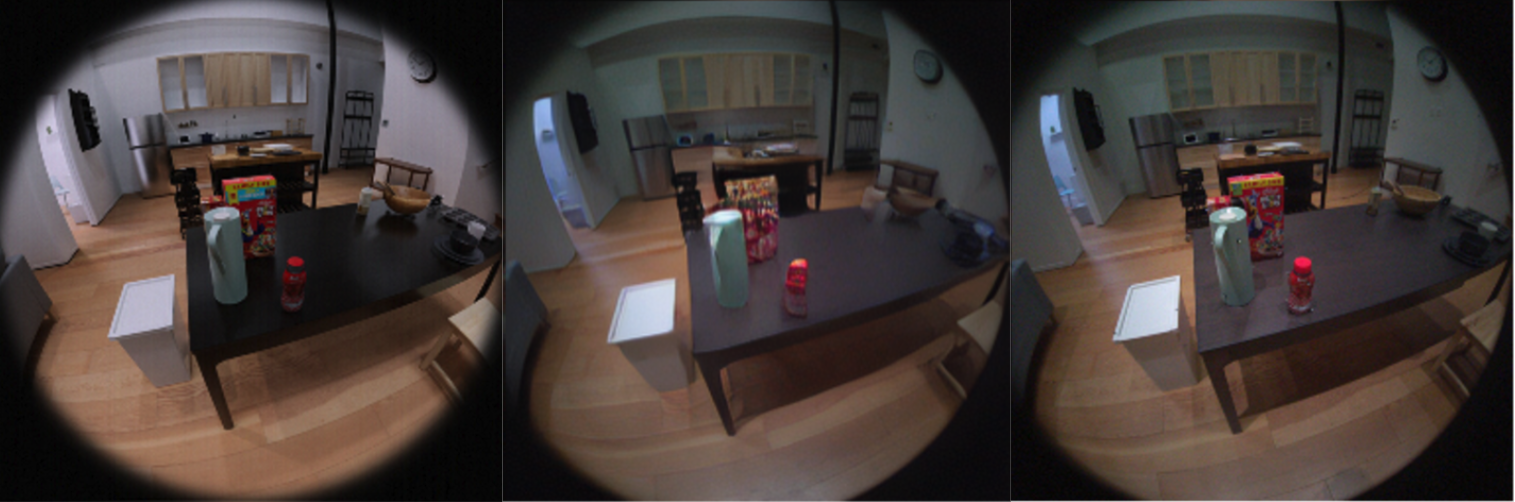}
\end{center}
\caption{An example of the domain transfer task. From left to right: synthetic RGB (source), LDM, real RGB (target).  }
\label{fig:benchmark_iit}
\end{figure}

Given our capability of rendering a synthetic twin for each ADT sequence, we explore the opportunity of closing the synthetic-to-real domain gap using image to image translation methods. We use ADT synthetic-real paired images to train four state-of-the-art methods; Pix2Pix~\cite{pix2pix2017}, TSIT~\cite{jiang2020tsit},  and LDM~\cite{rombach2022high}. The methods are trained on 43 sequences and evaluated on 102 unseen sequences. We benchmark the synthetic to real image translation performance by quantifying pixel-level distance with peak signal-to-noise ratio (PSNR), structural similarity (SSIM)~\cite{wang2004image} metrics, and a perceptual-level distance metric with the perceptual similarity metric (LPIPS)~\cite{zhang2018unreasonable}. Results are presented in Table~\ref{table:benchmarking_iit}, while Figure~\ref{fig:benchmark_iit} shows the qualitative results of an example frame. 

\begin{table}[!t]
\begin{center}
\begin{tabular}{|c|c|c|c|c|c|}
\hline
      & chair & bed & table  & fridge & sofa \\ \hline
Cube R-CNN & 3.72 & 2.947 & 2.796 & 2.601      & 1.252             \\ \hline
Total3D & 0.847 & 0.630 & 2.228 & 0.048          & 1.298          \\ \hline
\end{tabular}
\end{center}
\caption{AP (in \%) of top 5 categories for the 3D object detection task.}
\label{table:benchmarking_3d}
\end{table}

\begin{table}[!t]
\begin{center}
\begin{tabular}{|c|c|c|c|}
\hline
        & PSNR $\uparrow$           & SSIM $\uparrow$           & LPIPS $\downarrow$         \\ \hline
SyntheticADT & 16.383          & 0.456          & 0.270          \\ \hline
Pix2Pix & 23.442          & \textbf{0.674} & 0.162          \\ \hline
TSIT    & 21.885          & 0.617          & 0.161          \\ \hline
LDM     & \textbf{24.218} & 0.660          & \textbf{0.126} \\ \hline
\end{tabular}
\end{center}
\caption{Image-to-Image translation benchmarking.}
\label{table:benchmarking_iit}
\end{table}

\section{Conclusion}
\label{section:conclusion}
We introduced ADT, the most comprehensive egocentric dataset available to date. ADT includes 200 sequences captured with the sensor-rich Aria glasses in two fully digitized spaces: an apartment and an office. We described the state-of-the-art digitization process used to achieve photo-realism allowing for synthetic-real twins of each sequence. We described the precise ground truth generation procedure for object/Aria 6DoF poses, human poses and eye gazing, with an in-depth analysis of the total system accuracy. We then demonstrated the usefulness of the dataset by benchmarking important AR-related machine perception tasks including object detection, segmentation, and image translation. Overall, ADT pushes the boundaries of high quality, comprehensive egocentric datasets, unlocking new research opportunities for the community that would not have been possible previously.

{\small
\bibliographystyle{ieee_fullname}
\bibliography{egbib}
}

\clearpage
\newpage

\section{Supplementary Material}
In this section, we dive deep into the implementation of the system accuracy measurement and more detailed results of it. We perform more qualitative and quantitative analyses on the 2D object detection, image segmentation and 3D object detection tasks. Furthermore, we introduce another important use case of the ADT dataset that can quantitatively evaluate a manual 3D bounding box annotation pipeline before it is applied to large-scale egocentric data.

\subsection{System Accuracy}

We provide additional information and figures in this section to better describe the methodology. We also provide additional tables with results for the reader to better understand the data statistics and how the accuracy of the system depends on different factors. 

\begin{figure}[!t]
\begin{center}
  \includegraphics[width=0.9\linewidth]{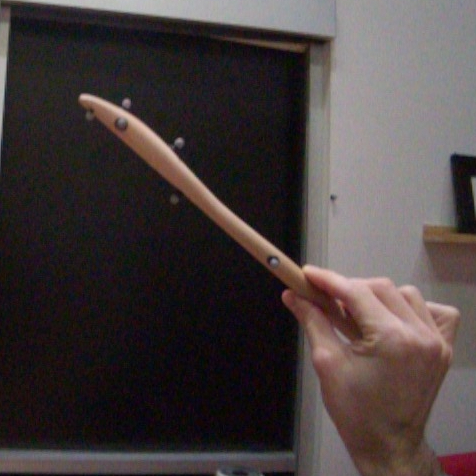}
\end{center}
\caption{Cropped version of example Aria image used for system accuracy tests.}
\label{fig:sup:error_quant_ini2}
\end{figure}

\begin{figure}[!t]
\begin{center}
  \includegraphics[width=0.9\linewidth]{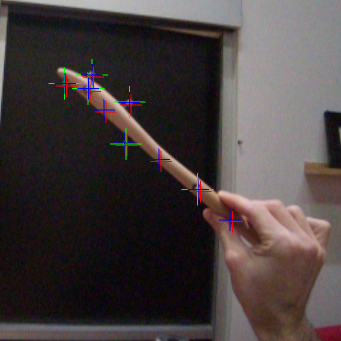}
\end{center}
\caption{Cropped version of example Aria image used for system accuracy tests with results. Red: system's estimate of where the markers should project. Green: hand labels of where the markers are located in the image. Blue: system estimate of where the markers should be after optimizing for the true object relative pose.}
\label{fig:sup:error_quant_ini3}
\end{figure}

Figures \ref{fig:sup:error_quant_ini2}  and \ref{fig:sup:error_quant_ini3} illustrate the system accuracy analysis on an exemplar frame. Figure \ref{fig:sup:error_quant_ini2} shows a portion of a zoomed in RGB image where a wooden spoon is being moved in by an Aria wearer. As described in Section 3.3, we take this image and manually label the centers of each marker. The system accuracy estimation pipeline then estimated the object pose relative to the image which best aligns the projection of the 3D markers to the hand labels. Figure \ref{fig:sup:error_quant_ini3} shows the final results after the optimization described in Section 3.3. The green crosses are the manual labels; the red crosses are the marker reprojections onto the image plane given all system measurements at the capture time for this frame; and the blue crosses are the reprojections of markers after applying the optimized object pose using Eqn.5 in Section 3.3. The misalignment between the green crosses and the red crosses indicates the error of the object pose. The alignment between the green crosses and the blue crosses confirms that the estimation of the true object poses is correct. 




Table \ref{table:sys_acc_results_scenes} shows the system accuracy statistics for each of the two scenes. The accuracy in the office is slightly better than the accuracy in the Apartment. We expect the root cause to be the higher ceilings in the apartment, where the motion capture cameras are installed, yielding a slightly worse tracking accuracy. Table \ref{table:sys_acc_results_objects} shows the system accuracy measurement of 32 dynamic objects averaged on a per-object basis. The total system error comes from the 3D object reconstruction, motion capture system, Aria device poses and Aria device calibration.

\begin{table*}[t]
    \centering
    \begin{tabular}{|l|c |c |c |c|}
    \hline
        Object Name & Measurement   & Translation & Rotation    & Reprojection    \\
                    & Count         & Error [mm]  & Error [deg] & Error[pixel]    \\ \hline
        BlackCeramicBowl & 10 & 3.05 & 0.66 & 5.05   \\ \hline
        Donut\_B & 11 & 3.61 & 1.06 & 4.84   \\ \hline
        MuffinPan & 10 & 3.64 & 0.59 & 5.45   \\ \hline
        RedClock & 10 & 3.72 & 1.03 & 4.19   \\ \hline
        DecorativeBoxHexLarge & 12 & 3.77 & 1.05 & 5.05   \\  \hline
        CoffeeCan\_2 & 10 & 4.06 & 0.66 & 5.43   \\  \hline
        Mortar & 11 & 4.19 & 0.74 & 6.45  \\  \hline
        ChoppingBoard & 10 & 4.25 & 0.49 & 5.23   \\ \hline
        BlackCeramicDishLarge & 10 & 4.31 & 0.71 & 5.26   \\ \hline
        WoodenFork & 13 & 4.53 & 1.65 & 6.71  \\  \hline
        BirdhouseToy\_2 & 17 & 4.77 & 1.11 & 4.55  \\ \hline
        BambooPlate & 10 & 4.82 & 0.67 & 7.34  \\  \hline
        BirdHouseToy & 12 & 5.08 & 0.72 & 7.53   \\  \hline
        Orange\_A & 14 & 5.22 & 2.28 & 8.19   \\  \hline
        ToothBrushHolder & 12 & 5.32 & 1.66 & 7.24   \\  \hline
        CakeMocha\_A & 15 & 5.62 & 0.69 & 6.14   \\  \hline
        WoodenSpoon & 10 & 5.85 & 2.02 & 6.42  \\ \hline
        WoodenBowl & 10 & 5.85 & 0.74 & 6.53   \\  \hline
        BlackPictureFrame & 13 & 6.00 & 1.16 & 8.73   \\  \hline
        BlackTablet & 7 & 6.19 & 1.11 & 6.69   \\  \hline
        BlackCeramicMug & 10 & 6.53 & 1.69 & 6.59   \\  \hline
        BookDeepLearning & 11 & 6.56 & 0.96 & 10.31   \\  \hline
        WoodenBoxSmall & 12 & 6.73 & 1.28 & 8.83  \\  \hline
        Flask & 14 & 7.17 & 1.49 & 5.71   \\  \hline
        GreenDecorationTall & 10 & 8.02 & 1.37 & 8.81   \\  \hline
        BlackRoundTable & 11 & 8.43 & 0.65 & 5.72   \\  \hline
        Cracker & 10 & 8.49 & 2.25 & 7.20  \\  \hline
        BlackKitchenChair & 9 & 12.24 & 0.79 & 5.66   \\  \hline
        WhiteChair & 6 & 12.35 & 0.77 & 6.83  \\  \hline
        Jam & 14 & 12.57 & 1.52 & 7.32   \\  \hline
        Cereal & 9 & 16.29 & 2.18 & 11.82  \\  \hline
        DinoToy & 10 & 25.39 & 4.65 & 7.25   \\ \hline
    \end{tabular}
    \caption{Mean system accuracy results for select objects ranked by the translation error.}
    \label{table:sys_acc_results_objects}
\end{table*}

\begin{table}[h]
\begin{tabular}{|c|c |c|}
\hline
Error  & Apartment  & Office  \\ \hline
Object translation [mm]  & 6.94 & 4.48 \\  \hline
Object rotation [deg]  & 1.3 & 1.04 \\  \hline
Reprojection Measured [pixels] & 6.9 & 4.18 \\  \hline
Reprojection Optimized [pixels] & 0.56 & 0.47 \\ \hline
\end{tabular}
\caption{Mean system accuracy results, split by scene location.}
\label{table:sys_acc_results_scenes}
\end{table}

\subsection{Performance Analysis on 2D Object Detection and Image Segmentation}
\label{section:sup:2d}
\begin{figure*}[!t]
\begin{center}
  \includegraphics[width=0.9\linewidth]{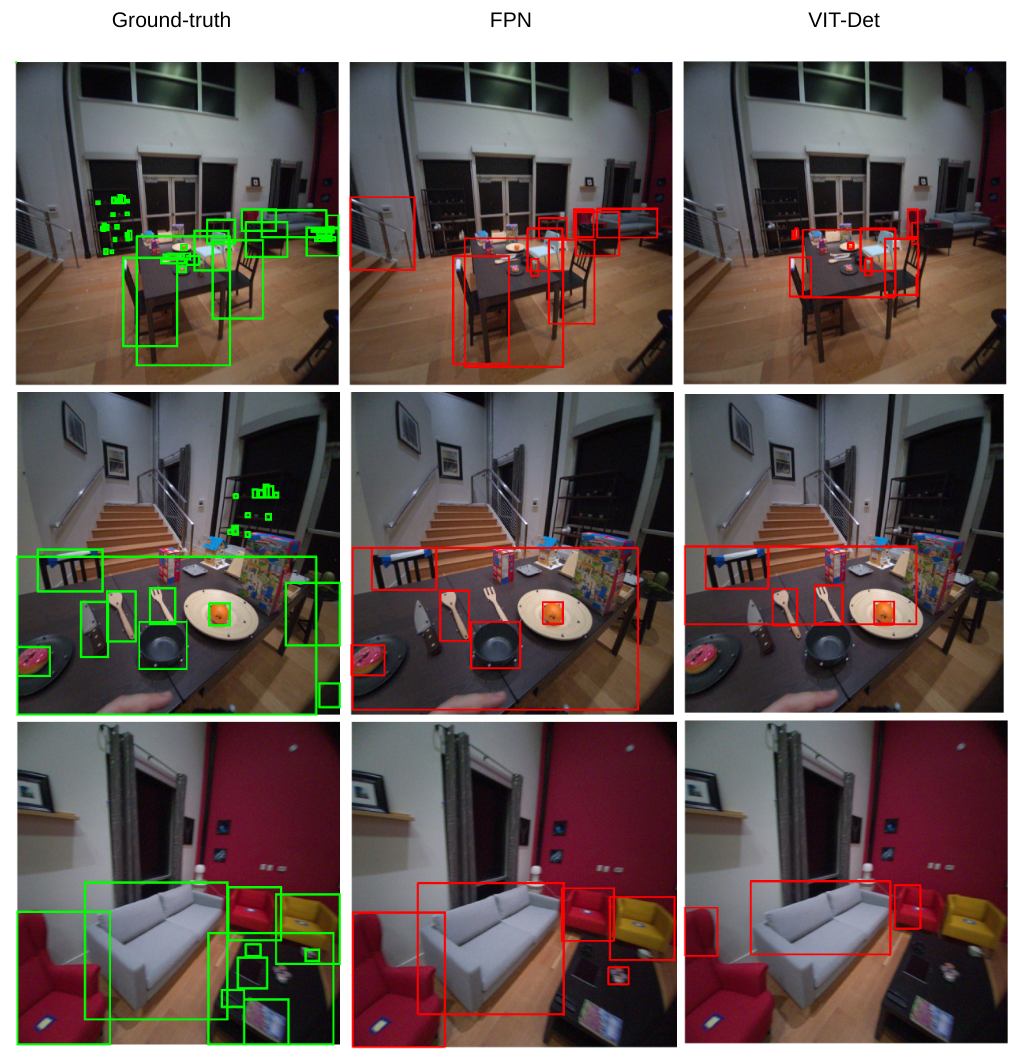}
\end{center}
\caption{Each row is an example of the comparison among the ground-truth, FPN 2D detection result and VIT-Det 2D detection result. All three examples shows that FPN tends to detect larger objects better than that of VIT-Det, such as the dining table in the first and second example, and the sofa and armchairs in the third example. FPN also shows promising robustness results under view point variance such as the dining table in the second example and the leftmost armchair in the third example. In contrast, VIT-Det seems to be better at detecting smaller objects such as the bottles on the shelf behind the dining table in the first example and the fork in the second example.}
\label{fig:sup:2d_comparison}
\end{figure*}

The performance of the state-of-the-art models, namely FPN and VIT-Det, for 2D object detection and image segmentation tasks on the ADT dataset is significantly lower than their performance on the COCO dataset. We expect this discrepancy is largely due to the domain difference between these two datasets, which is consistent with the findings of~\cite{TREK150ijcv}. Despite the rectification of the Aria fisheye RGB images to bring ADT closer to the distribution of COCO, the egocentric nature of the data still remains a challenge for these algorithms. Table~\ref{table:2d_per_category_results} shows the per-category mAP. As can be seen from the table, large furniture, appliances categories such as couch, chair, refrigerator are typically easier for the detectors to detect in these videos while their performance is poor on object categories such as potted plant, mouse, remote etc. Though this can be attributed to the scale of the objects present in the videos, it also highlights the challenges of building a real world index of everyday objects from in the wild recordings. Furthermore, in a qualitative analysis, Figure ~\ref{fig:sup:2d_comparison} show the performance of both detectors along with the ground truth. FPN shows better performance detecting large objects and objects under viewpoint variance. Although VIT-Det seems to be better at detecting small objects compared to FPN, its overall inferior performance to FPN suggests a possible mismatch between the training scale and the sizes of the ADT images at the inference stage.  

\begin{table}[t]
    \centering
    \begin{tabular}{|l| c | c | c | c |}
    \hline
        Category & FPN & FPN     & VIT-Det &VIT-Det \\ \hline
                 & Box & Seg  & Box    & Seg \\ \hline
        Frisbee & 18.55 & 21.10 & 7.51 & 6.80 \\ \hline
        Bottle & 2.91 & 3.03 & 1.28 & 1.32 \\ \hline
        Cup & 5.67 & 5.64 & 4.56 & 4.83 \\ \hline
        Fork & 8.12 & 2.85 & 4.25 & 1.13 \\ \hline
        Knife & 14.50 & 10.58 & 10.82 & 7.93 \\ \hline
        Spoon & 14.20 & 6.24 & 7.07 & 3.78 \\ \hline
        Bowl & 17.81 & 17.41 & 7.23 & 7.53 \\ \hline
        Banana & 16.87 & 12.73 & 8.25 & 6.32 \\ \hline
        Apple & 21.64 & 24.03 & 12.31 & 14.03 \\ \hline
        Sandwich & 14.15 & 10.94 & 8.88 & 11.41 \\ \hline
        Orange & 19.84 & 21.80 & 9.87 & 10.80 \\ \hline
        Carrot & 37.08 & 53.02 & 38.84 & 29.75 \\ \hline
        Donut & 3.93 & 4.57 & 2.29 & 2.54 \\ \hline
        Cake & 10.25 & 12.52 & 9.21 & 10.84 \\ \hline
        Chair & 34.38 & 17.44 & 20.80 & 9.58 \\ \hline
        Couch & 49.77 & 49.87 & 27.82 & 32.20 \\ \hline
        Potted Plant & 0.51 & 0.48 & 0.40 & 0.38 \\ \hline
        Bed & 7.29 & 2.42 & 6.34 & 3.61 \\ \hline
        Dining Table & 25.02 & 7.63 & 2.37 & 0.75 \\ \hline
        TV & 24.73 & 29.65 & 19.10 & 23.76 \\ \hline
        Laptop & 12.66 & 12.78 & 2.30 & 2.61 \\ \hline
        Mouse & 1.11 & 0.98 & 0.20 & 0.17 \\ \hline
        Remote & 1.47 & 0.30 & 1.82 & 0.54 \\ \hline
        Keyboard & 4.01 & 3.31 & 0.44 & 0.30 \\ \hline
        Oven & 0.05 & 0.01 & 0.61 & 0.37 \\ \hline
        Toaster & 0.09 & 0.11 & 2.22 & 2.54 \\ \hline
        Refrigerator & 48.47 & 48.45 & 42.89 & 43.63 \\ \hline
        Book & 10.12 & 9.23 & 3.40 & 2.83 \\ \hline
        Clock & 34.33 & 34.97 & 32.21 & 33.37 \\ \hline
        Vase & 0.34 & 0.28 & 0.22 & 0.12 \\ \hline
        Scissors & 7.52 & 0.14 & 10.92 & 0.33 \\ \hline

    \end{tabular}
    \caption{Per-category 2D detection and segmentation mean mAP computed across all videos in the dataset. Large furniture and appliances are easier to detect for the detectors than the smaller objects like remotes. This indicates the challenges in the constructing real world index of everyday objects.}
    \label{table:2d_per_category_results}
\end{table}

\subsection{Performance Analysis of 3D Object Detection}
\begin{figure*}[t]
\begin{center}
  \includegraphics[width=0.9 \linewidth]{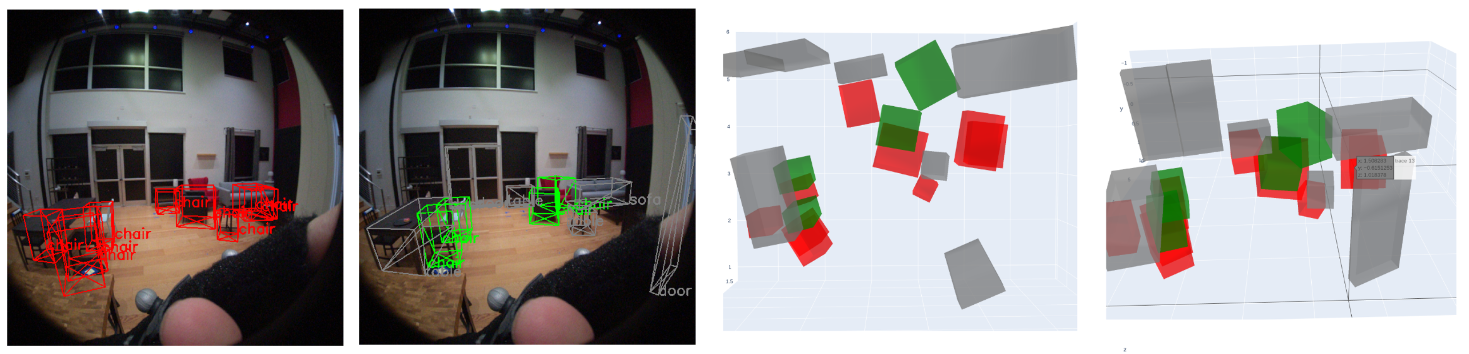}
  \subcaption{A failure example of Cube R-CNN on predicting 3D poses of chairs.}
  \label{fig:3d_detection_failure_chair}
  \includegraphics[width=0.9 \linewidth]{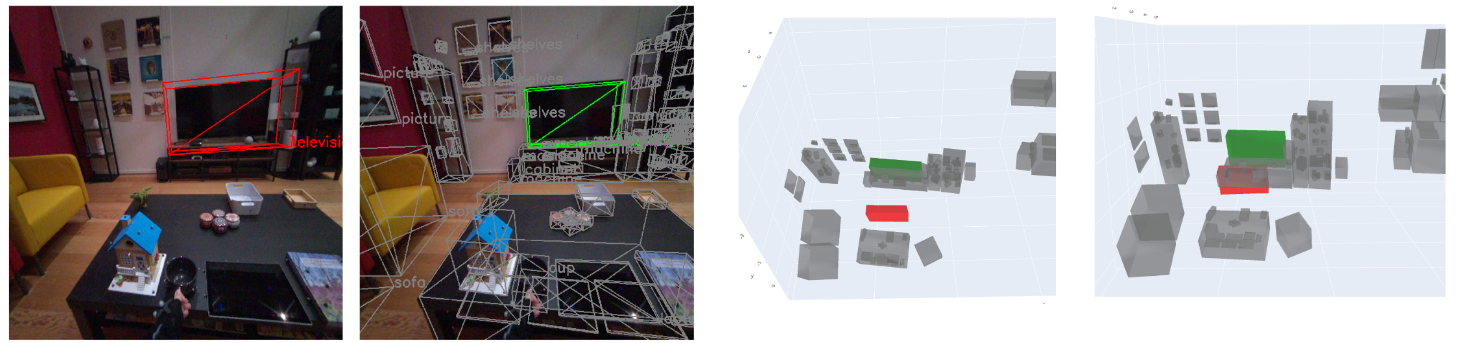}
  \subcaption{A failure example of Total3d on predicting the 3D pose of a TV object.}
  \label{fig:3d_detection_failure_tv}
  \includegraphics[width=0.9 \linewidth]{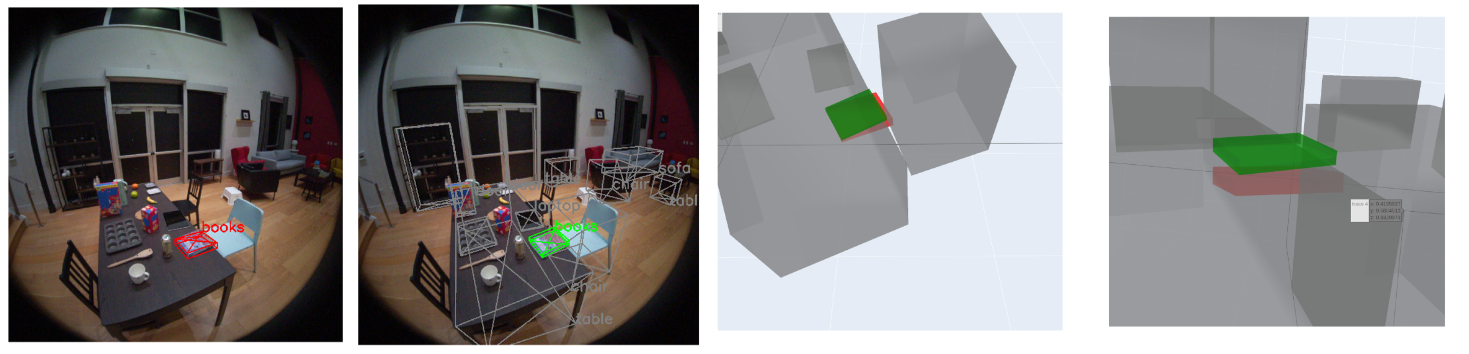}
  \subcaption{A failure example of Total3d on predicting the 3D pose of a book object.}
  \label{fig:3d_detection_failure_book}
\end{center}
\caption{From left to right: 3D object detection in red bounding boxes; ground truth bounding boxes  in green for the target object and in gray for other objects; predicted 3D bounding boxes from a top down view; predicted 3D bounding boxes from a side view. }
\label{fig:manual_annotation_examples}
\end{figure*}

The 3D object detection performance of Cube-RCNN and Total3d is significantly lower on the ADT dataset. We therefore conduct more analyses on the failure cases to enlighten the challenges of 3D object detection research. Our observations include two major failure cases: 1) 2D object detection failure, 2) 3D pose prediction failure. Since we analyse 2D object detection failures in Section \ref{section:sup:2d}, we will focus on 3D pose prediction failures in this section. Figure~\ref{fig:3d_detection_failure_chair} shows a typical failure case of 3D pose prediction. Cube R-CNN roughly localizes the 3D position of eight chairs but fails in predicting 3D poses accurately enough to pass the IoU threshold of 0.25. 

Additionally, we observed frequent failure cases with the depth estimation which is a fundamental limitation of 3D detection models based on single image inputs, since 3D data is challenging to infer from a single 2D image. Figure~\ref{fig:3d_detection_failure_tv} and Figure~\ref{fig:3d_detection_failure_book} show two failure examples for Total3D and Cube R-CNN, respectively. The reprojected 3D bounding boxes fit well on the 2D images. However as evident from the 3D visualizations, the predicted poses are significantly erroneous when compared to the ground truth. This problem can be potentially solved by a more advanced 3D object detector using multi-camera sensors from Aria. 

\subsection{Comparison with Manual 3D Bounding Box Annotation}
Accurate 3D bounding boxes in the ADT ground truth dataset can be leveraged to benchmark the accuracy of a video-based manual annotation pipeline. To set up the evaluation, we select 20 randomly sampled videos (10\% of the total videos) from the dataset for manual annotation of 3D bounding boxes using objects from 10 categories. Figure~\ref{fig:manual_annotation_examples} shows examples of the manual annotations. We evaluate each manual bounding box annotation of an object by computing the difference from the 6DoF ground truth pose in ADT, including translation, rotation and scale errors. The mean translation error is 0.329 meters; the mean rotation error is 4.29 deg and the mean relative scale error is 0.32. We show the evaluation results on three example categories in Table~\ref{table:benchmarking_3d_manual}.

The experiment above introduces a distinct advantage for testing a semi-automatic annotation pipeline and for training annotators with continuous, quantified and visualized feedback before creating large-scale tasks. Visualizations such as those shown in Figure~\ref{fig:manual_annotation_examples} can act as a quick reference for educating annotation teams on the common failure modes and patterns.

\begin{table}[t]
\begin{center}
\begin{tabular}{|c|c|c|c|}
\hline
 & Sofa & Photo Frame & Chair\\ \hline
Center Prediction (m) &  0.296  & 0.162 & 0.041 \\ \hline
Rotation (deg) &  3.869 & 1.952 & 1.553 \\ \hline
Relative Scale & 0.15 & 0.27 & 0.10\\ \hline
\end{tabular}
\end{center}
\caption{Benchmarking of the manual annotations. It shows error in manually annotated objects measured against the accurate ground truth provided by the ADT. Smaller objects are difficult to annotate with accuracy as can be seen from the higher relative scale error of the photo frames.}
\label{table:benchmarking_3d_manual}
\end{table}

\begin{figure*}[t]
\begin{center}
  \includegraphics[width=0.25 \linewidth]{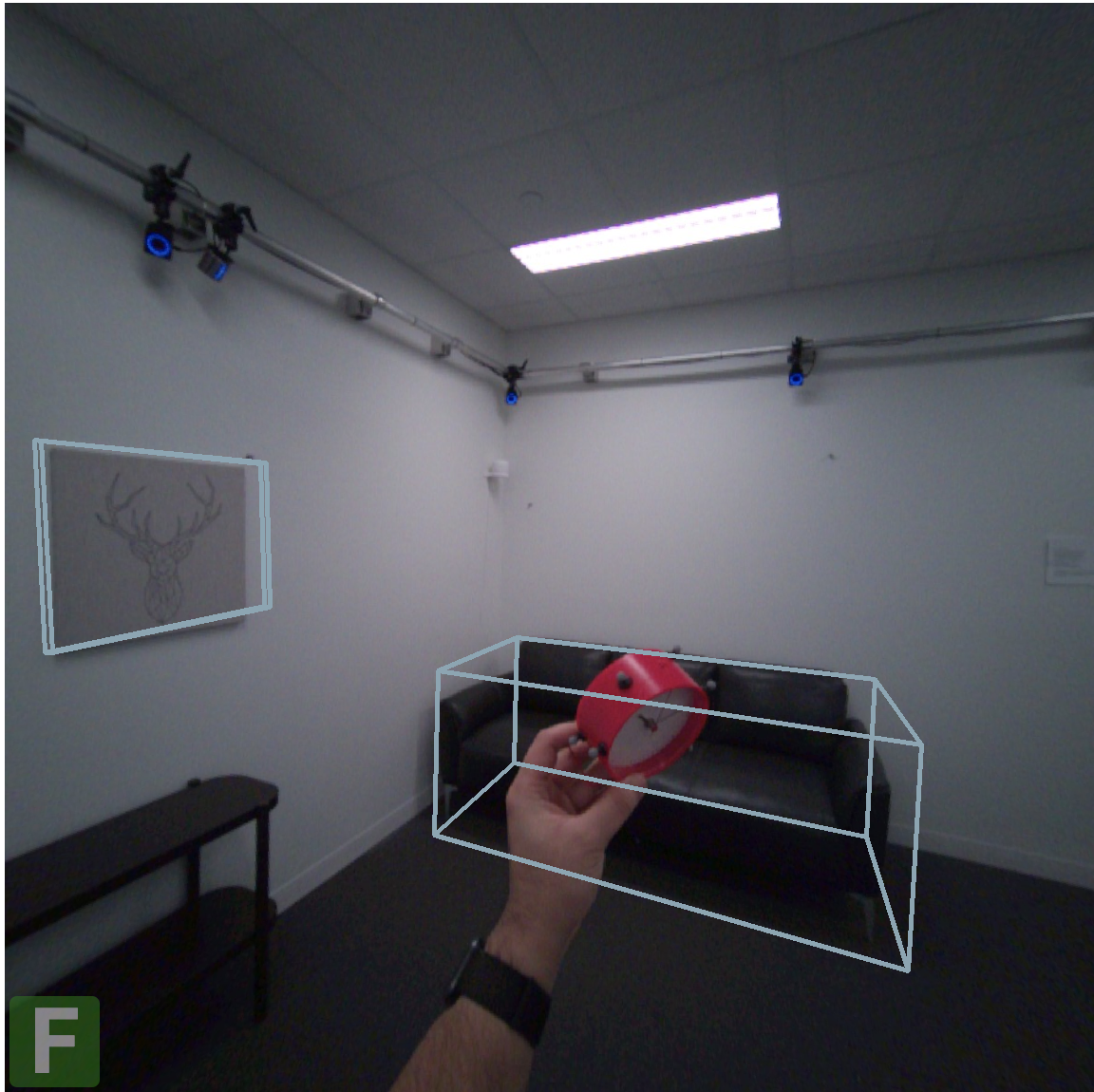}
  \includegraphics[width=0.25 \linewidth]{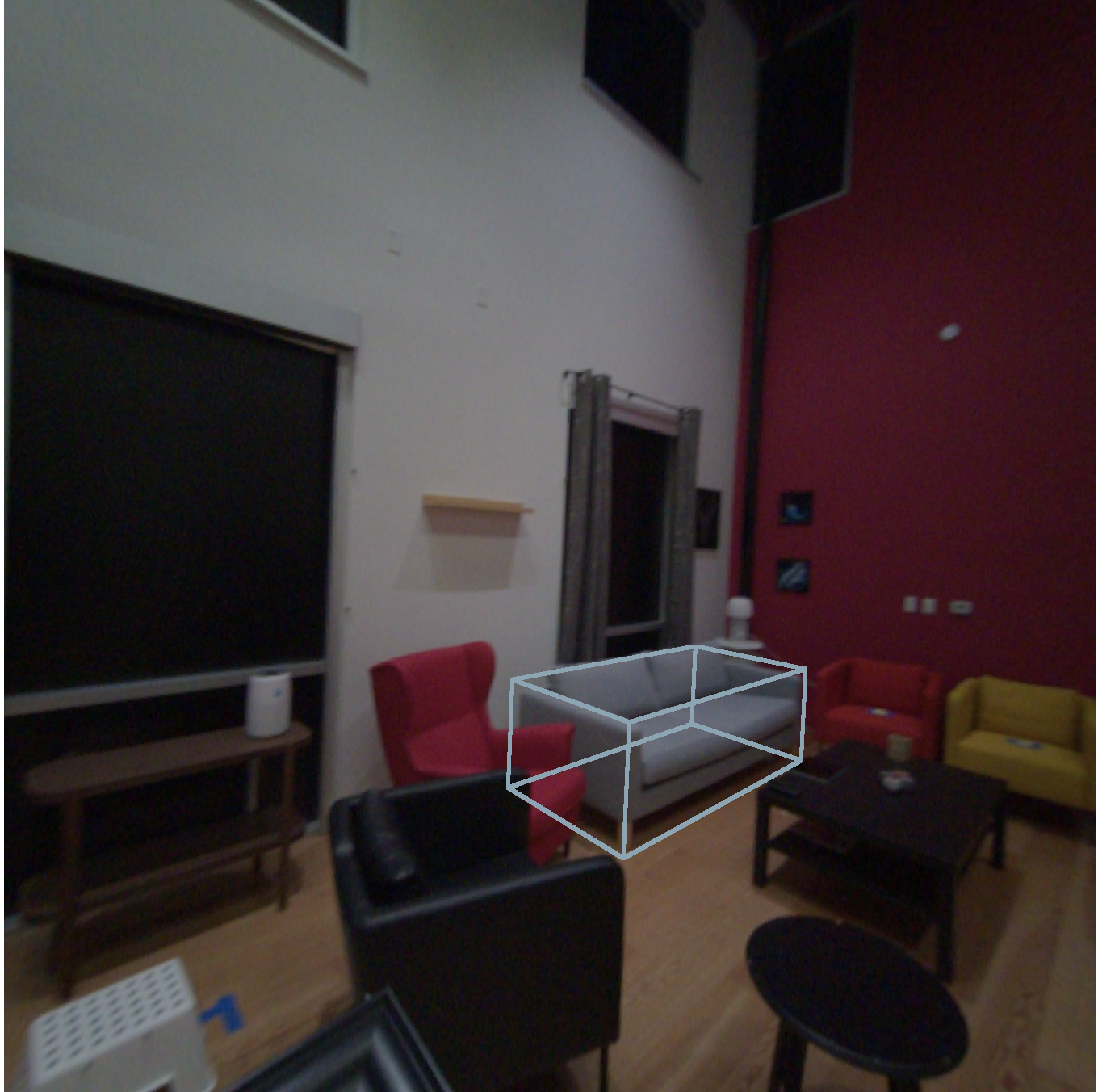}
  \includegraphics[width=0.25 \linewidth]{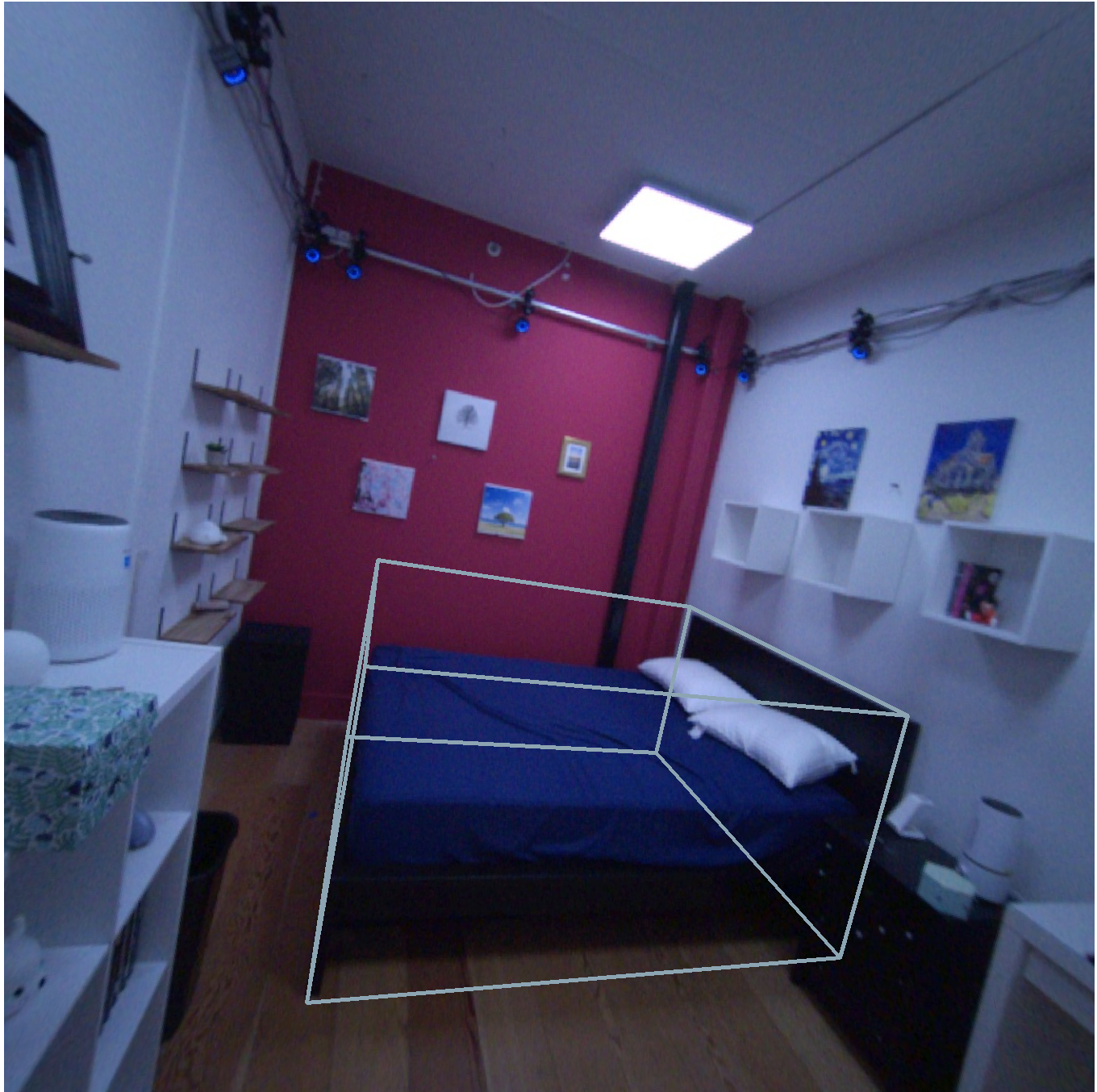}
\end{center}
\caption{Examples of the manual annotation. Small and thin objects are typically more difficult to manually annotate compared to large and bulky objects. The error margin for annotating a photo frame is much smaller as compared to annotating bigger furniture objects such as the sofa and bed. Typically annotating the depth becomes a challenging task and is often the main cause of the error. The ADT dataset allows for an accurate estimate of these errors as shown in table~\ref{table:benchmarking_3d_manual} }
\label{fig:manual_annotation_examples}
\end{figure*}

\end{document}